%% file: wacv19_main.tex
\documentclass[10pt,twocolumn,letterpaper]{article}

\usepackage{wacv}
\usepackage{times}
\usepackage{epsfig}
\usepackage{graphicx}
\usepackage{amsmath}
\usepackage{amssymb}

\usepackage{comment}
\usepackage{float}
\usepackage{booktabs}
\usepackage{caption}
\usepackage{afterpage}
\usepackage{adjustbox}
\usepackage{enumitem}

\def\mytabspace{\vspace{-3mm}}

\def\mytabcapspace{\vspace{-3mm}}

\wacvfinalcopy 

\def\wacvPaperID{408} 

\newcommand{\Paragraph}[1]{\vspace{1.25mm} \noindent \textbf{#1} \hspace{0mm}}

\ifwacvfinal\fi
\begin{document}


\title{Recurrent Flow-Guided Semantic Forecasting} 

\author{Adam M. Terwilliger, Garrick Brazil, Xiaoming Liu\\
Department of Computer Science and Engineering\\
Michigan State University\\
{\tt\small \{adamtwig, brazilga, liuxm\}@msu.edu}
}


\twocolumn[{%
\renewcommand\twocolumn[1][]{#1}%
\maketitle
\begin{center}
    \centering
\includegraphics[width=0.88\linewidth]{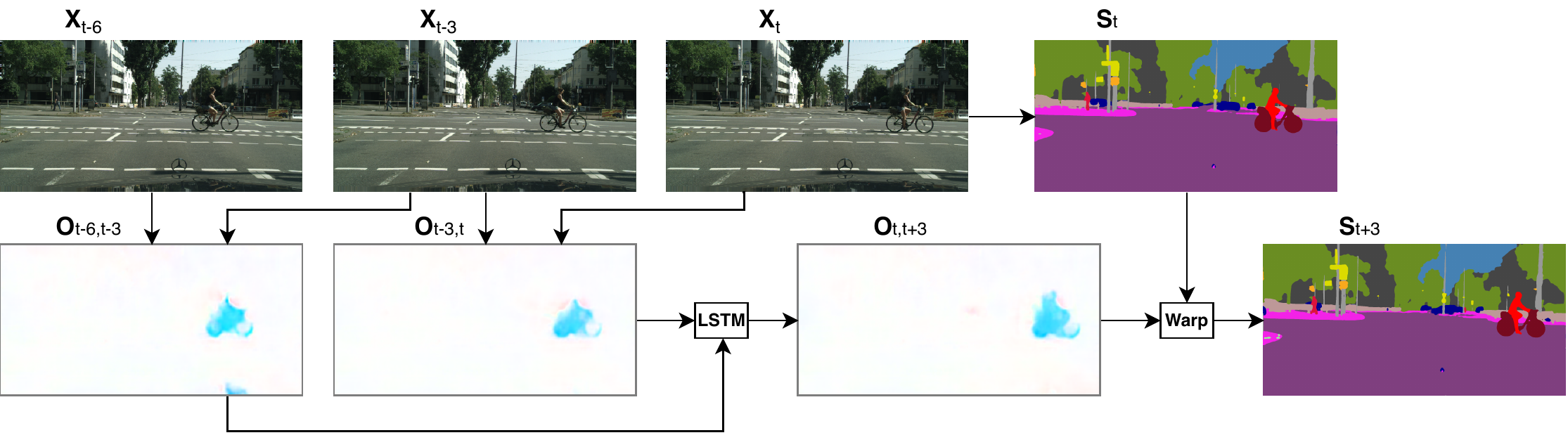}
\captionof{figure}{Our proposed approach aggregates past optical flow features using a convolutional LSTM to predict future optical flow, which is used by an learnable warp layer to produce future segmentation.}
\label{fig:model_viz}
\end{center}%
}]


\input{wacv19_abs.tex}

\input{wacv19_intro.tex}

\input{wacv19_related.tex}

\input{wacv19_method.tex}

\input{wacv19_exp.tex}

\input{wacv19_con.tex}

{\small
\bibliographystyle{ieee}
\bibliography{wacv19_ref}
}

\end{document}

%% file: wacv19_abs.tex
\begin{abstract}
Understanding the world around us and making decisions about the future is a critical component to human intelligence. 
As autonomous systems continue to develop, their ability to reason about the future will be the key to their success. 
Semantic anticipation is a relatively under-explored area for which autonomous vehicles could take advantage of (e.g., forecasting pedestrian trajectories). 
Motivated by the need for real-time prediction in autonomous systems, we propose to decompose the challenging semantic forecasting task into two subtasks: current frame segmentation and future optical flow prediction. 
Through this decomposition, we built an efficient, effective, low overhead
model with three main components: flow prediction network, feature-flow aggregation LSTM, and end-to-end learnable warp layer. 
Our proposed method achieves state-of-the-art accuracy on short-term and moving objects semantic forecasting while simultaneously reducing model parameters by up to $95\%$ and increasing efficiency by greater than $40x$. 
\end{abstract}

%% file: wacv19_intro.tex
\section{Introduction}
Reasoning about the future is a crucial element to the deployment of real-world computer vision systems. 
Specifically, garnering semantics about the future of an urban scene is essential for the success of autonomous driving. 
Future prediction tasks are often framed through video prediction (i.e., given $n$ past RGB frames of a video, infer about $m$ future frames). 
There is extensive prior work in directly modeling future RGB frames~\cite{Cricri_NIPS2016, Denton_NIPS2017, Finn_NIPS2016, Kalchbrenner_ICML2017, Lotter_ICLR2017, Mahjourian_IV2017}. 
Although RGB reconstruction may provide insights into representation learning, the task itself is exceedingly complex. 
On the other hand, real-time autonomous systems could make more intelligent decisions through semantic anticipation (e.g., forecasting pedestrian \cite{Brazil_ICCV2017} or vehicle \cite{Chen_CVPR2016_2}). 
Recent work has shown that decomposing complex systems into semantically meaningful components can improve performance and enable better long-term planning~\cite{Shalev_2_arXiv2016,Shalev_arXiv2016}.

Scene understanding through semantic segmentation is one of the most successful tasks in deep learning~\cite{LeCun_DL2, LeCun_DL1}. 
Most approaches focus on reasoning about the current frame \cite{Badrinarayanan_PAMI2017, Chen_arXiv2016, Long_CVPR2017, Yu_ICLR2016, Zhao_CVPR2017} through convolutional neural networks (CNNs). 
However, recently a new task was proposed by Luc \etal~\cite{Luc_ICCV2017} for prediction of future segmentation frames, termed \textit{semantic forcasting}. 
Luc \etal~\cite{Luc_ICCV2017} note that directly modeling pixel-level semantics instead of intermediate RGB provides a greater ability to model scene and object dynamics. 
They further propose a multi-scale auto-regressive CNN as a baseline for the task. 
Jin \etal~\cite{Jin_NIPS2017} improve performance even further using a multi-stage CNN to jointly predict segmentation and optical flow.

There are three limitations of prior approaches for semantic forecasting. 
First, both \cite{Jin_NIPS2017} and \cite{Luc_ICCV2017} use a deep CNN to both learn and apply a warping operation which transforms past segmentation features into the future.
However, Jaderberg \etal~\cite{Jaderberg_NIPS2015} have shown that despite the power of CNNs, they have limited capacity to spatially move and transform data purely through convolution. 
Hence, this inability extends to optical flow warping, an essential re-mapping operation, as exemplified in Fig.~\ref{fig:model_viz}. 
Therefore, it is generally easier for a CNN to estimate warp parameters than it is to both predict \textit{and} warp through convolution alone.
Secondly, Luc \etal~\cite{Luc_ICCV2017} and Jin \etal~\cite{Jin_NIPS2017} do not directly account for the inherent temporal dependency of video frames between one another. 
Rather, both approaches concatenate four past frames as input to a CNN which restricts the network to learn long-term (e.g., $> 4$ frames) dependencies. 
Finally, \cite{Jin_NIPS2017} and \cite{Luc_ICCV2017} were not designed with low overhead and efficiency in mind, both of which are essential for real-time autonomous systems.

Our approach addresses each of these limitations with a simple, compact network. 
By decomposing motion and semantics, we reduce the semantic forecasting task into two subtasks: current frame segmentation and future optical flow prediction. 
We utilize a learnable warp layer~\cite{Zhu_ICCV2017,Zhu_CVPR2017} to precisely encode the warp into our structure rather than forcing a CNN to unnaturally learn and apply it at once, hence addressing the first limitation. 
We additionally utilize the temporal coherency of concurrent video frames through a convolutional long short-term (LSTM) module~\cite{Hochreiter_LSTM, Shi_NIPS2015}. 
We use the LSTM module to aggregate optical flow features over long term time, thus addressing the second limitation. 
By decomposing the problem into simpler subtasks, we are able to completely remove the CNN previously dedicated to processing concatenated segmentation features from past frames as input. 
Doing so dramatically reduces the number of parameters of our model and redistributes the bulk of the work to a lightweight flow network.
Such processing also efficiently uses single frame segmentation features and avoids redundant re-reprocessing of frames (first-in-first-out concat.). 
Hence, the decomposition leads to an efficient, low overhead model, and addresses the final limitation.

Our work can be summarized in three main contributions, novel with respect to the semantic forecasting task:
\begin{enumerate}[noitemsep,topsep=1mm]
\setlength\itemsep{2mm}
\item A learnable warp layer directly applied to semantic segmentation features,
\item A convolutional LSTM to aggregate optical flow features and estimate future optical flow,
\item A lightweight, modular baseline which greatly reduces the number of model parameters, improves overall efficiency, and achieves state-of-the-art performance on short-term and moving objects semantic forecasting.
\end{enumerate}

%% file: wacv19_related.tex
\section{Related Work}
Our work is largely driven by the decomposition of motion and semantics. 
Our underlying assumption is that the semantic forecasting task can be solved using \textit{perfect} current frame segmentation and a \textit{perfect} optical flow which warps current to future frames. 
Our underlying motivation for the decomposition is also found in~\cite{Villegas_ICLR2017}, which focuses on the future RGB prediction task. 
Further, \cite{Cheng_ICCV2017,Gadde_CVPR2017,Jin_ICCV2017} decompose motion and semantics, but focus on improving current frame segmentation, rather than on future frames. 

We review prior work with respect to our three main contributions: learnable warp layer, recurrent motion prediction, and lightweight semantic forecasting.

\Paragraph{Learnable warp layer.} 
Jaderberg \etal~\cite{Jaderberg_NIPS2015} provide a foundation for giving CNNs the power to learn transformations which are otherwise exceedingly difficult for convolution and non-linear activation layers to simultaneously learn and apply. 
Zhu \etal~\cite{Zhu_ICCV2017,Zhu_CVPR2017} utilize a differentiatable spatial warping layer which enables efficient end-to-end training. 
Ilg \etal~\cite{Ilg_CVPR2017} use a similar warping layer to enable stacks of FlowNet modules which iteratively improve optical flow.
Likewise, there are numerous other works which utilize a learnable warp layer~\cite{Jourabloo_CVPR2017, Liang_ICCV2017, Liu_ICCV2017, Lu_CVPR2017, Tran_CVPR2018}. 
Although \cite{Jin_NIPS2017} warps the RGB and segmentation features, they do not do so using an end-to-end warp layer. 
As such, the learnable warp layer in our model facilitates strong performance using a single prediction, rather than relying on multiple recursive steps. 
This influence is emphasized in our state-of-the-art results on short-term forecasting.

\begin{figure*}[t!]
\center
\includegraphics[width=0.73\linewidth]{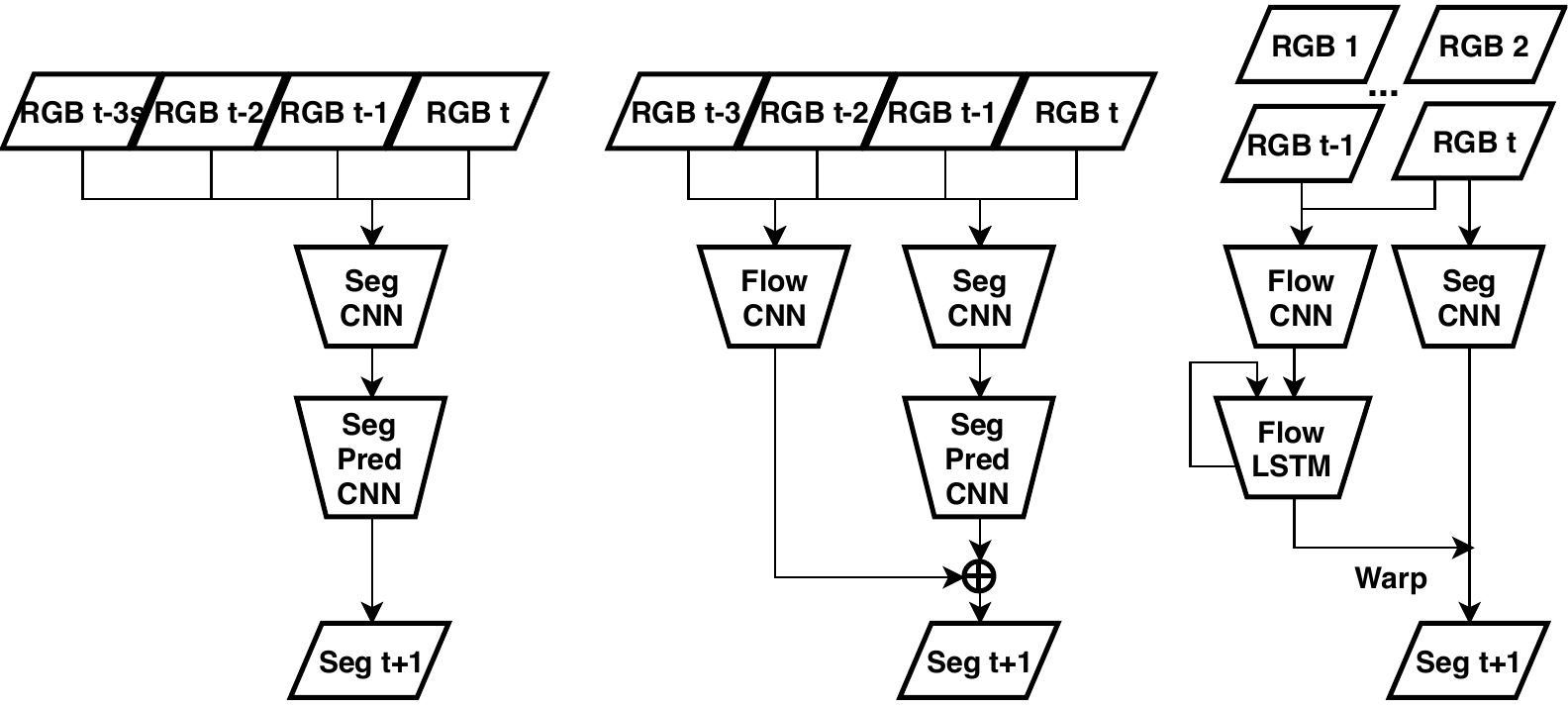}
\caption{The overall network architectures of Luc \etal~\cite{Luc_ICCV2017} vs. Jin \etal~\cite{Jin_NIPS2017} vs.~ours (from left to right).}
\label{fig:model_comp}
\end{figure*}

\Paragraph{Recurrent motion prediction.} 
Numerous works have been published in motion prediction~\cite{Alahi_CVPR2016, Chao_CVPR2017, Luo_CVPR2017, Morteza_BMVC2014, Villegas_ICML2017, Vondrick_CVPR2016,  Walker_ECCV2016, Walker_CVPR2014, Walker_ICCV2015}, which motivate flow prediction and using a recurrent structure for future anticipation. 
Villegas \etal~ \cite{Villegas_ICLR2017} encode motion as the subtraction between two past frames and utilize a ConvLSTM~\cite{Hochreiter_LSTM, Shi_NIPS2015} to better aggregate temporal features. 
Similar to our work, Patraucean \etal~\cite{Patraucean_ICLR2016} utilize an LSTM to predict optical flow that is used to warp segmentation features. 
However, \cite{Patraucean_ICLR2016} use a single RGB frame as input in their network, whereas our network utilizes a pair of frames, as seen in Fig.~\ref{fig:model_comp}. 
Further, we utilize the power of the FlowCNN to encode strong optical flow features as input to the LSTM, where \cite{Patraucean_ICLR2016} uses only a single convolutional encoding layer. 
Lastly, \cite{Patraucean_ICLR2016} utilize the future RGB frame to produce their segmentation mask, which does not enable their structure for semantic forecasting. 
As such, our method, to the best of our knowledge, is the first to aggregate optical flow features using a recurrent network and predict future optical flow for semantic forecasting.

\Paragraph{Semantic forecasting.} 
Fig.~\ref{fig:model_comp} compares the network structures of the two main baselines for semantic forecasting. 
Luc \etal~\cite{Luc_ICCV2017} formulates a two-stage network which first obtains segmentation features through the Dilation10 \cite{Yu_ICLR2016} current frame network (SegCNN). 
The four past segmentation features are then concatenated as input to their SegPredCNN, denoted $S_2S$, to produce the future segmentation frame. 
When doing mid-term prediction, $S_2S$ uses an auto-regressive approach, passing their previous predictions back into the network. 
The main differences between $S_2S$ and our proposed method include the removal of the SegPredCNN and addition of the FlowCNN. 
By directly encoding motion as a flow prediction, our method can achieve better overall accuracy on moving objects than~\cite{Luc_ICCV2017}.

Jin \etal~\cite{Jin_NIPS2017} improves upon $S_2S$ through its addition of a FlowCNN and joint training on both flow and segmentation. 
Note that both \cite{Luc_ICCV2017} and \cite{Jin_NIPS2017} utilize strictly up to $4$ past frames as input, while our method uses an LSTM which can leverage any number of past frames in the long-term.
Further, \cite{Jin_NIPS2017} utilize an extra convolutional layer to augment the predictions through a residual connection between SegPredCNN and FlowCNN.
All three CNNs in~\cite{Jin_NIPS2017} are based on Res101-FCN networks, which add numerous layers to ResNet-101~\cite{He_CVPR2016}.
The overarching motivation for \cite{Jin_NIPS2017} compared to our method is quite different. 
Jin \etal~emphasized the benefit of the flow prediction influencing semantic forecasting and vice versa. 
In contrast, we demonstrate the value of treating these tasks {\it independently}. 
Further, there are three main differentiating factors between \cite{Jin_NIPS2017} and ours: warp layer, FlowLSTM, and network size. 
With the combination of FlowLSTM and warp layer, we are able to forgo the SegPredCNN, thus dramatically reducing our network size, while maintaining high-fidelity segmentation results.

%% file: wacv19_method.tex
\section{Method}
Semantic forecasting is defined as the task when given $t$ input frames ${\bf X}_{1:t}$, predict the pixel-wise semantic segmentation for future frame ${\bf S}_{t+s}$, where $s$ is the future step. 
In this section, we will present models for both short-term $s=\{1,3\}$ and mid-term $s=\{9,10\}$ prediction.

\subsection{Short-term Prediction}
\subsubsection{Model Overview}
Our approach is motivated by the notion that semantic forecasting can be decomposed into two sub-tasks: current frame segmentation and future optical flow prediction. 
Specifically, we propose
\begin{equation}
{\bf S}_{t+s} = {\bf O}_{t\rightarrow t+s}({\bf S}_t),
\label{eqn:overallmodel}
\end{equation}
where ${\bf S}_t$ is current segmentation and ${\bf O}_{t\rightarrow t+s}$ is the ground truth optical flow which defines the warp of ${\bf S}_t$ into ${\bf S}_{t+s}$. 
With much recent attention focused on current frame segmentation \cite{Badrinarayanan_PAMI2017, Chen_arXiv2016, Long_CVPR2017, Yu_ICLR2016, Zhao_CVPR2017}, our model is designed to be agnostic to the choice of backbone segmentation network. 
Thus, we focus on estimating future optical flow $\hat{{\bf O}}_{t\rightarrow t+s}$. 
Our method for this estimation contains three components: FlowCNN, FlowLSTM, and warp layer. 
The FlowCNN is designed to extract high-level optical flow features. 
The FlowLSTM aggregates flow features over time and produces a future optical flow prediction, while maintaining long-term memory of past frame pairs. 
The warp layer uses a learnable warp operation which applies the flow prediction directly to the segmentation features produced by the segmentation network. 
Since the warp layer is learnable, we utilize backpropagation through time (BPTT) for efficient end-to-end training of all three components. 

\subsubsection{FlowCNN}
Although most state-of-the-art optical flow methods are non-deep learning, a critical caveat is that such approaches run on CPU,  which can take anywhere between $5$ min to an hour for a single frame. 
Since runtime efficiency is a priority for our work, we utilize FlowNet \cite{Fischer_2015,Ilg_CVPR2017} for optical flow estimation with a CNN.
Specifically, we use the FlowNet-c architecture with pre-trained weights from FlowNet2~\cite{Ilg_CVPR2017}. 
Utilizing the pre-trained weights is important because no ground truth optical flow is provided in Cityscapes~\cite{Cordts_Cityscapes}. 
The FlowCNN takes a concatenated pair of RGB frames as input and is denoted as: 
\begin{equation}
\hat{{\bf O}}_{t-s\rightarrow t} = f_{F}({\bf X}_{t-s}, {\bf X}_{t}).
\label{eqn:FlowCNN}
\end{equation}

Our reasoning for using FlowNet-c is twofold: simple and effective. 
Firstly, this network has only approximately $5$ million parameters, which consumes less than $2$ GB of GPU memory at test time and can evaluate in less than $1$ second. 
Secondly, although Jin \etal~\cite{Jin_NIPS2017} train their FlowCNN from scratch using weak optical flow ground truths, we note that through utilization of FlowNet-c our method demonstrates improved short-term performance with a dramatic reduction in parameters. 
However, we emphasize that our method is not fixed to only FlowNet-c.
Instead it is compatible with any optical flow network, since the flow features can be extracted as input into the FlowLSTM, thus demonstrating the modularity of our method.  
Since our method is not fixed to a specific SegCNN or FlowCNN architecture, as performance on these individual tasks improve over time, our method can improve with them, demonstrating its potential longevity as a general-purpose baseline for semantic forecasting. 

\subsubsection{FlowLSTM} 
One of the main differentiating factors between our proposed method and previous semantic forecasting work is how time is treated. 
Specifically, both \cite{Luc_ICCV2017} and \cite{Jin_NIPS2017} do not use a recurrent structure to model the inherent temporal dynamics of past frames. 
Instead, $4$ past frames are concatenated thereby ignoring temporal dependency among frames $>4$ time-steps away from $t$ and encodes spatial redundancy. 
In contrast, by utilizing a LSTM module, our method can retain long-term memory as frames are processed over time. 

When a pair of input frames is passed through the FlowCNN, the flow features are extracted at a certain level of the refinement layers. 
We observe similar performance regardless of the level of the refinement layer extracted (predict\_conv6--predict\_conv2), with one exception. 
We find it less effective to {directly} aggregate the flow field predictions compared to aggregating the flow features. 

As such, we extract flow features immediately before the final prediction layer of FlowNet-c, which represent $74$ channels at $128\times256$ resolution. 
We use these features as input to the ConvLSTM with a $3\times 3$ kernel and $1$ padding. 
We then produce the future optical flow prediction using a single convolution layer with $1 \times 1$ kernel after the ConvLSTM to reduce the channels from $74 \to 2$ for the flow field. 
The FlowLSTM can be defined as:
\begin{equation}
\hat{{\bf O}}_{t\rightarrow t+s} = f_{L}(\hat{{\bf O}}_{1\rightarrow s+1},\hat{{\bf O}}_{2\rightarrow s+2},..., \hat{{\bf O}}_{t-s\rightarrow t}).
\end{equation}

\subsubsection{Warp Layer}
Recall that FlowCNN extracts motion from the input frames and FlowLSTM aggregates motion features over time. 
Thus, the natural next question is how to use these features for the semantic forecasting task? 
Both Luc \etal~\cite{Luc_ICCV2017} and Jin \etal~\cite{Jin_NIPS2017} utilize a SegPredCNN which attempts to extract motion from the segmentation features \textit{and} apply the motion {simultaneously}.
Jin \etal~\cite{Jin_NIPS2017} combines the features extracted from the FlowCNN with features extracted from the SegPredCNN through a residual block which learns a weighted summation. 
By giving a CNN the ability to directly apply the warp operation with a warp layer rather than through convolutional operations, we can forgo the SegPredCNN that both \cite{Luc_ICCV2017} and \cite{Jin_NIPS2017} rely heavily upon.
To define the warp layer, we must first define the SegCNN:
\begin{equation}
{\bf \hat{S}}_{t} = f_{S}({\bf X_t}).
\label{eqn:SegCNN}
\end{equation}
Therefore, using equations~\ref{eqn:FlowCNN}--\ref{eqn:SegCNN}, the warp layer can be formally defined as:
\begin{equation}
{\bf \hat{S}}_{t+s} = f_W(f_{L}(f_{F}({\bf X}_1, {\bf X}_{s+1}),..., f_{F}({\bf X}_{t-s}, {\bf X}_{t})), f_{S}({\bf X_t})).
\label{eqn:warp}
\end{equation}
Specifically, the warp layer takes a feature map with an arbitrary number of channels as input.
Then using a continuous variant of bilinear interpolation, we compute the warped image by re-mapping w.r.t.~the flow vectors. 
We define all pixels to be zero where the flow points outside of the image.

Finally, we reduce equation~\ref{eqn:warp} as an approximation for the \textit{perfect} warp function seen in equation~\ref{eqn:overallmodel}:
\begin{equation}
{\bf \hat{S}}_{t+s} = f_W({\bf \hat{O}}_{t\rightarrow t+s}, {\bf \hat{S}}_t).
\end{equation}

\subsubsection{Loss function}

For short-term prediction, the main loss function used is cross-entropy loss with respect to the ground truth segmentation for the $20$th frame. 
We note that due to the recurrent nature of our method, we do not need to rely on weak segmentation $L_{\ell _1 }$ or $L_{\text{gdl}}$ gradient difference loss of \cite{Luc_ICCV2017,Jin_NIPS2017}. 
Thus our main loss is formulated as:
\begin{equation}
L_{\text{seg}} ({\bf\hat S_{t+s}},{\bf S_{t+s}}) =  - \sum\limits_{(i,j) \in {\bf X_t} } {{y_{i,j} \log p({\bf\hat S}_{t+s}^{i,j}}) },
\end{equation}
where $y_{i,j}$ is the ground truth class for the pixel at location $(i,j)$ and $p({\bf \hat S}_{t+s}^{i,j})$ is the predicted probability that ${\bf \hat S}_{t+s}$ at location $(i,j)$ is class $y$.

\subsection{Mid-term Prediction}
There are two main types of approach for mid-term prediction: single-step and auto-regressive. \cite{Luc_ICCV2017} and \cite{Jin_NIPS2017} respectively define mid-term prediction as being nine and ten frames forward, approximately $0.5$ seconds in the future. 
The single-step approach uses past frames ${\bf X}_{1:t}$ to predict a single optical flow ${\bf \hat{O}}_{t\rightarrow t+m}$, where $m$ is a mid-term jump $(9,10)$, to warp ${\bf S}_{t}$ to ${\bf S}_{t+m}$. 
In contrast, the auto-regressive approach takes multiple steps to reach ${\bf S}_{t+m}$, warping both the segmentation features and the last RGB frame to pass back into the FlowCNN as input for the next step.

For example, consider when $m=10$. For single-step, our model estimates the optical flow ${\bf\hat{O}}_{10\rightarrow 20}$ and warps ${\bf S}_{10} \to \hat{S}_{20}$. 
For the auto-regressive approach, our model instead estimates the shorter optical flow ${\bf\hat{O}}_{10\rightarrow 15}$ and warps ${\bf S}_{10} \to {\bf \hat{S}}_{15}$ and ${\bf X}_{10} \to {\bf \hat{X}}_{15}$. 
Then ${\bf X}_{10}$ and ${\bf \hat{X}}_{15}$ are passed back as input into the FlowCNN to produce the optical flow ${\bf\hat{O}}_{15\rightarrow 20}$ which warps ${\bf\hat{S}}_{15} \to {\bf\hat{S}}_{20}$.

\begin{figure}[t!]
\centering
\includegraphics[width=\linewidth]{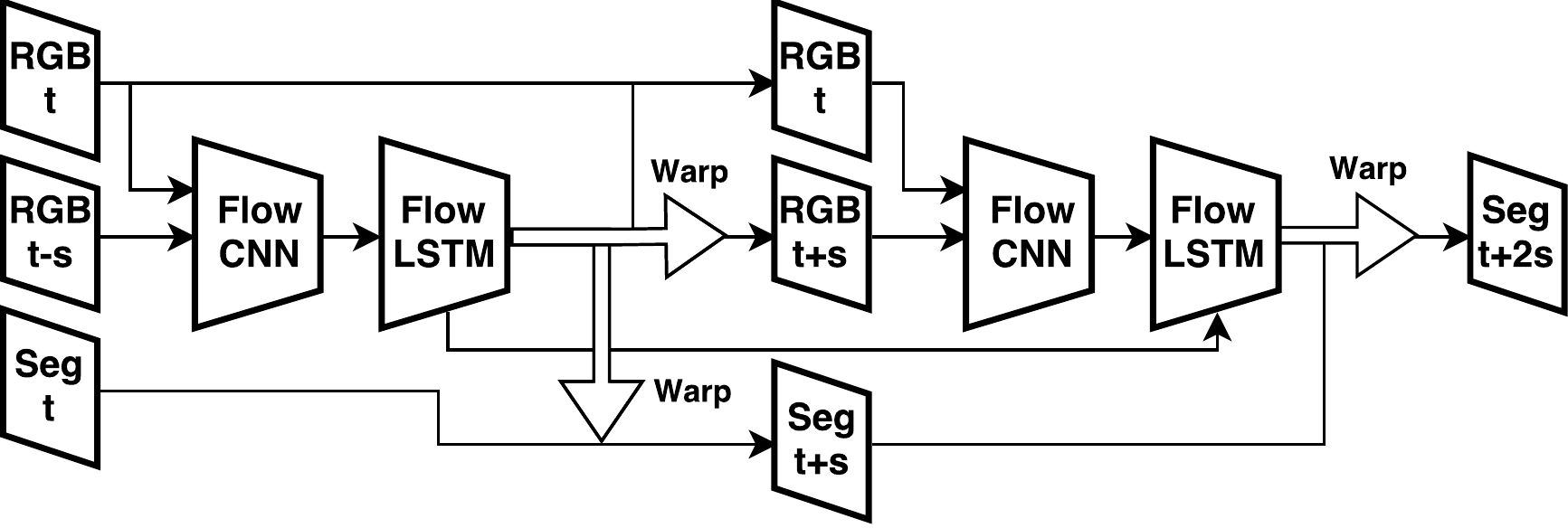}
\caption{Unrolled network structure for mid-term prediction via auto-regressive approach.}
\label{fig:unroll}
\end{figure}

The benefits of the single-step approach are more efficient training / testing and no error propagation through regressive use its own \textit{noisy} prediction. 
In contrast, the auto-regressive approach has the benefit of being a single model for any time step, as well as, being able to regress indefinitely into the future. 
Additionally, most optical flow methods are designed for next frame or short-term flow estimation. 
Therefore, a single large step may not be as easy to estimate compared to multiple small steps. 
Further, both approaches can learn non-linear flow due to the presence of the FlowLSTM.
For the auto-regressive approach, the unrolled recurrent structure can be visualized in Fig.~\ref{fig:unroll}. 

\subsubsection{Loss function}
For mid-term prediction, we utilize not only the cross-entropy loss but also an RGB reconstruction loss. 
Specifically for the auto-regressive approach, with each small step, we take the $\ell _1$ loss between warped ${\bf \hat{X}}$ and ground truth ${\bf X}$. 
For more stable training, we use the {\it smooth} $l_1$ loss~\cite{Girshick_ICCV2015}, by:
\begin{equation}
\small
  \begin{aligned}
    \label{eq:loss_seg}
    L_{\text{rgb}} ({\bf\hat X_{t+s}},{\bf X_{t+s}}) = 
    \begin{cases}
      0.5(d_{\ell _1 })^2 , & |d_{\ell _1 }| < 1, \\
     |d_{\ell _1 }| - 0.5, & \text{otherwise}
    \end{cases}
  \end{aligned}
\end{equation}
where $d_{\ell _1 }$ is the $\ell _1$ distance as:
$$d_{\ell _1 }({\bf\hat X_{t+s}},{\bf X_{t+s}}) = \sum\limits_{(i,j) \in {\bf X_{t+s}} } {\left| {{\bf\hat X_{t+s}}^{i,j}  - {\bf X_{t+s}}^{i,j} } \right|}.$$

The full loss function is
\begin{equation}
L = \lambda_1 L_{\text{seg}} + \lambda_2 L_{\text{rgb}},
\end{equation}
where $\lambda_1$ and $\lambda_2$ are weighting factors treated as hyper-parameters during the training process. 
We find setting $\lambda_1 = \lambda_2 = 1$ early on during training and then iteratively lowering $\lambda_2$ closer to 0 provides optimal results.

%% file: wacv19_exp.tex
\section{Experiments}

\subsection{Dataset and evaluation criteria}

Our experiments are focused on the urban street scene dataset Cityscapes~\cite{Cordts_Cityscapes}, which contains $5{,}000$ high-quality images ($1{,}024 \times 2{,}048$) with pixel-wise annotations for $19$ semantic classes (pedestrian, car, road, etc.). 
Further, $19$ temporal frames preceding each annotated semantic segmentation frame are available. 
Our method is able to utilize all $19$ previous frames for the $2{,}975$ training sequences.

We report performance of our models on the $500$ validation sequences using the standard metric of mean Intersection over Union (IoU) \cite{Everingham_IoU}:
\begin{equation*}
IoU = \frac{TP}{TP+FP+FN},
\end{equation*}
where TP, FP, and FN represent the numbers of true positive, false positive, and false negative pixels, respectively. We compute IoU with respect to the ground truth segmentation of the $20$th frame in each sequence. 
To further demonstrate the effectiveness of our method on moving objects, we compute the mean IoU on the $8$ available foreground classes (\emph{person, rider, car, truck, bus, train, motorcycle, and bicycle}) denoted as IoU-MO.

\begin{table*}
\centering
{\setlength{\tabcolsep}{0.95em}
\begin{tabular}{lccccc} \toprule &   Training & Testing & Params & Training & Testing\\ 
Model   & (hours) & (sec) & (mil) & GPUs & Method\\ 
    \midrule
    S$_2$S \cite{Luc_ICCV2017} & $96$ $(8x)$& $0.248$ $(4.8x)$& {$ \bf \approx$$1.5$} ($+300\%$) & $1$ & Single \\
    SP-MD \cite{Jin_NIPS2017} & -- & $2.182$ $(42x)$ & $\approx$$115$ ($-95\%$) & $4$-$8$ & Sliding \\
    \midrule
       ours & $\bf{12}$ & $\bf{0.052}$ & $\approx$$6.0$ & $1$ & Single \\
    \bottomrule
  \end{tabular}}
 \caption{Computational complexity analysis with respect to previous work -- Luc \etal~\cite{Luc_ICCV2017} and Jin \etal~\cite{Jin_NIPS2017}. Models are measured without the SegCNN included (only SegPred). Runtime estimates were calculated by averaging $100$ forward passes with each model. Single vs.~sliding testing describes using a single forward pass of $512 \times 1,024$ resolution, relative to the costly sliding window approach of eight overlapping $713 \times 713$ full resolution crops.
 }
    \label{tab:comp}
\end{table*}

\begin{table}[t!]
\begin{center}
{\setlength{\tabcolsep}{0.3em}
\begin{tabular}{lcccc}
\toprule
 & IoU & IoU-MO & IoU & IoU-MO \\
Model & $(t=3)$ & $(t=3)$ & $(t=9)$ & $(t=9)$\\
\midrule
Copy last input \cite{Luc_ICCV2017} & $49.4$ & $43.4$ & $36.9$ & $26.8$ \\
Warp last input \cite{Luc_ICCV2017} & $59.0$ & $54.4$ & $44.3$ & $37.0$  \\
 S$_2$S \cite{Luc_ICCV2017} & $59.4$ & $55.3$ & $47.8$ & $40.8$ \\
\midrule
ours & $\bf67.1$ & $\bf65.1$ & $\bf51.5$ &  $\bf46.3$ \\
\bottomrule
\end{tabular}}
\end{center}
      \mytabcapspace
 \caption{Comparison of baselines with Luc \etal~\cite{Luc_ICCV2017} for short-term ($t=3$) and mid-term ($t=9$) for all nineteen classes and eight foreground, moving objects (MO) classes.}
   \label{tab:compLuc}
\end{table}

\subsection{Implementation details}
In our experiments, the resolution of any input image, ${\bf X}$, to either the SegCNN or FlowCNN is set to $512 \times 1{,}024$. 
The segmentation features, ${\bf \hat{S}}$, extracted from the SegCNN are also at $512 \times 1{,}024$, while those extracted from PSP-half are at $64 \times 128$. 
The optical flow prediction, ${\bf \hat{O}}$, from FlowNet-c is produced at $128 \times 256$. 
In most experiments, we upsample both ${\bf \hat{S}}$ and ${\bf \hat{O}}$ to $512 \times 1{,}024$. 
For training hyperparameters, we trained with SGD using the poly learning rate policy with initial an learning rate of $0.001$, power set to $0.9$, weight decay set to $0.0005$, and momentum set to $0.9$. 
Additionally, we used a batch size of $1$ and  trained for $50$k iterations ($\approx$$17$ Epochs). 
All of our experiments were conducted using either a NVIDIA Titan
X or a $1080$ Ti GPU and using the Caffe~\cite{caffe} framework.

\begin{table}[t!]
\begin{center}
{\setlength{\tabcolsep}{0.95em}
\begin{tabular}{lcc} \toprule Model  & IoU ($t=1$) & IoU ($t=10$)  \\
    \midrule
    Copy last input \cite{Jin_NIPS2017} & $59.7$ & $41.3$ \\
    Warp last input \cite{Jin_NIPS2017} & $61.3$ & $42.0$ \\
    SP-MD$^{\dagger}$ \cite{Jin_NIPS2017} & -- & $52.6$ \\
    SP-MD \cite{Jin_NIPS2017} & $66.1$ & $\bf{53.9}$ \\
        \midrule
     ours (c)$^{\dagger}$ & $73.0$ & $51.8$ \\
     ours (C) $^{\dagger}$ & $\bf{73.2}$ & $52.5$ \\
    \bottomrule
  \end{tabular}}
  \end{center}
 \caption{Comparison of baselines with Jin \etal~\cite{Jin_NIPS2017} for short-term ($t=1$) and mid-term ($t=10$). $\dagger$ - model contained no recurrent fine-tuning.  We compare our models with FlowNet2-c and FlowNet2-C backbones, where C contains $\frac{8}{3}$ more feature channels ($24$ vs.~$64$, $48$ vs.~$128$, etc.).}
    \label{tab:nips}
\end{table}

\subsection{Baseline comparison}

We compare our method to the two main previous works in semantic forecasting on Cityscapes: \cite{Luc_ICCV2017} and \cite{Jin_NIPS2017}. 
Our comparison is multifaceted: number of model parameters (in millions), model efficiency (in seconds), and model performance (in mean IoU). 
Additionally, we compare against \cite{Luc_ICCV2017} on the more challenging foreground moving objects (IoU-MO) baselines. 
To more directly compare with \cite{Luc_ICCV2017}, we utilize their definition of short-term as $t=3$ and mid-term as $t=9$. However, \cite{Jin_NIPS2017} redefines short-term as $t=1$ and mid-term as $t=10$. 
Thus, without reproducing each of their methods, we distinctly compare against each definition of short-term and mid-term prediction separately. 

\subsubsection{Baselines:}
\Paragraph{Copy last input:} \cite{Luc_ICCV2017,Jin_NIPS2017} use current frame segmentation, ${\bf\hat{S}_{t}}$, as prediction for future frame segmentation, ${\bf\hat{S}_{t+s}}$.

\Paragraph{Warp last input:} \cite{Jin_NIPS2017, Luc_ICCV2017} warp current frame segmentation, ${\bf\hat{S}_{t}}$, using future optical flow, ${\bf\hat{O}_{t\rightarrow t+s}}$. 
\cite{Luc_ICCV2017} estimates future optical flow using FullFlow \cite{Chen_CVPR2016} with the underlying assumption that ${\bf\hat{O}_{t\rightarrow t+s}} \approx$$-{\bf\hat{O}_{t\rightarrow t-s}}$. 
\cite{Jin_NIPS2017} estimates the optical flow with a FlowCNN (Res101-FCN) trained from scratch using EpicFlow \cite{Revaud_CVPR2015} for weak ground truths.

\Paragraph{S$_2$S:} Luc \etal~\cite{Luc_ICCV2017} generate $4$ previous frame segmentations: ${\bf\hat{S}_{t-3s}},{\bf\hat{S}_{t-2s}},{\bf\hat{S}_{t-s}},{\bf\hat{S}_{t}}$ using the current frame network from \cite{Yu_ICLR2016} then concatenate the features as input to the multi-scale CNN from \cite{Mathieu_ICLR2016} as their SegPredCNN.

\Paragraph{SP-MD.} Jin \etal~\cite{Jin_NIPS2017} also generate $4$ previous frame segmentations: ${\bf\hat{S}_{t-3s}}$,~${\bf\hat{S}_{t-2s}}$,~${\bf\hat{S}_{t-s}}$,~${\bf\hat{S}_{t}}$ using their own Res101-FCN network and concatenate as input to their SegPredCNN (Res101-FCN). Simultaneously, their FlowCNN (Res101-FCN) takes in $4$ previous RGB frames ${\bf X_{t-3s}}$, ${\bf X_{t-2s}}$,
${\bf X_{t-s}}$,${\bf X_{t}}$ to generate future optical flow ${\bf\hat{O}_{t\rightarrow t+s}}$. 
Lastly, features from the FlowCNN and SegPredCNN are fused via a residual block to make the final prediction. 

\begin{figure*}
\centering
\includegraphics[width=0.95\linewidth]{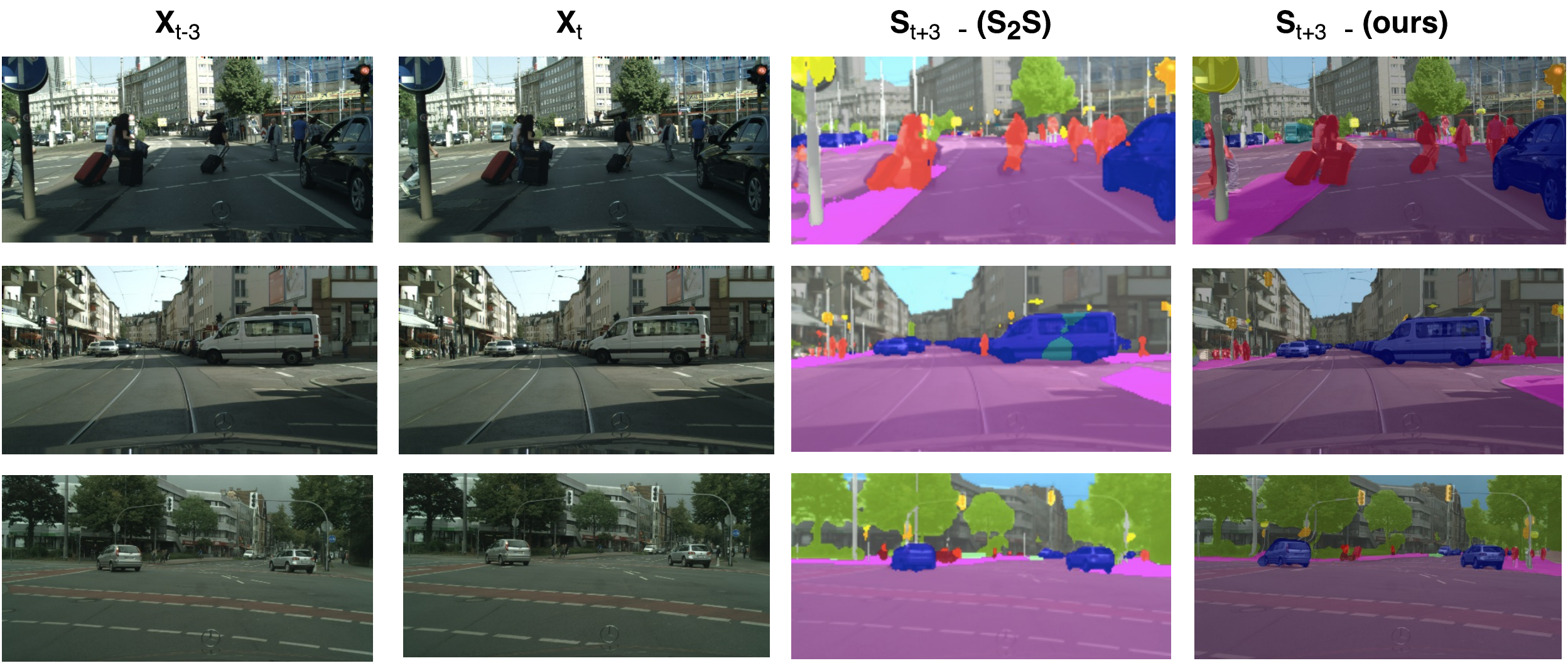}
\caption{
S$_2$S \cite{Luc_ICCV2017} vs. ours. Qualitative improvements are demonstrated, specifically with respect to moving objects such as pedestrian and vehicle trajectories.} 
\label{fig:short-term}
\end{figure*}

\subsubsection{Analysis}
The computational complexity of our methods relative to \cite{Luc_ICCV2017} and \cite{Jin_NIPS2017}  is in Tab.~\ref{tab:comp}. Since our work is motivated by deployment of real-time autonomous systems, we emphasize both low overhead and efficiency.  
Our approach demonstrates a significant decrease in the number of SegPred parameters relative to \cite{Jin_NIPS2017}. 
Specifically, we reduce the number of parameters from $\approx$$115M$ to $\approx$$6.0M$ resulting in a dramatic $95\%$ decrease. 
This large reduction can be directly linked to our removal of the large SegPredCNN replaced with the lightweight FlowNet-c as our FlowCNN combined with the warp layer. 
With respect to efficiency, our method sees a $4.8x$ and $42x$ speedup overall relative to \cite{Luc_ICCV2017} and \cite{Jin_NIPS2017}, respectively.
We infer this speedup is correlated with our need to only process a single past frame's segmentation features and our ability to limit the number of redundant image processing steps.

Although, we increase the number of SegPred parameters from $\approx$$1.5M$ to $\approx$$6.0M$ relative to \cite{Luc_ICCV2017}, we find an appreciable performance increase on short-term, mid-term, and moving objects prediction (Tab.~\ref{tab:compLuc}). Specifically, we improve by $7.7\%$ and $3.7\%$ on short-term and mid-term, respectively. On the more challenging moving objects benchmark, we improve on short-term by $9.8\%$ and mid-term by $5.5\%$. Since we directly encode motion into our model through our FlowCNN and aggregate optical flow features over time with our FlowLSTM, we can infer more accurate trajectories on moving objects, seen in Fig. \ref{fig:short-term}.

\begin{figure}
\center
\includegraphics[width=0.49\linewidth]{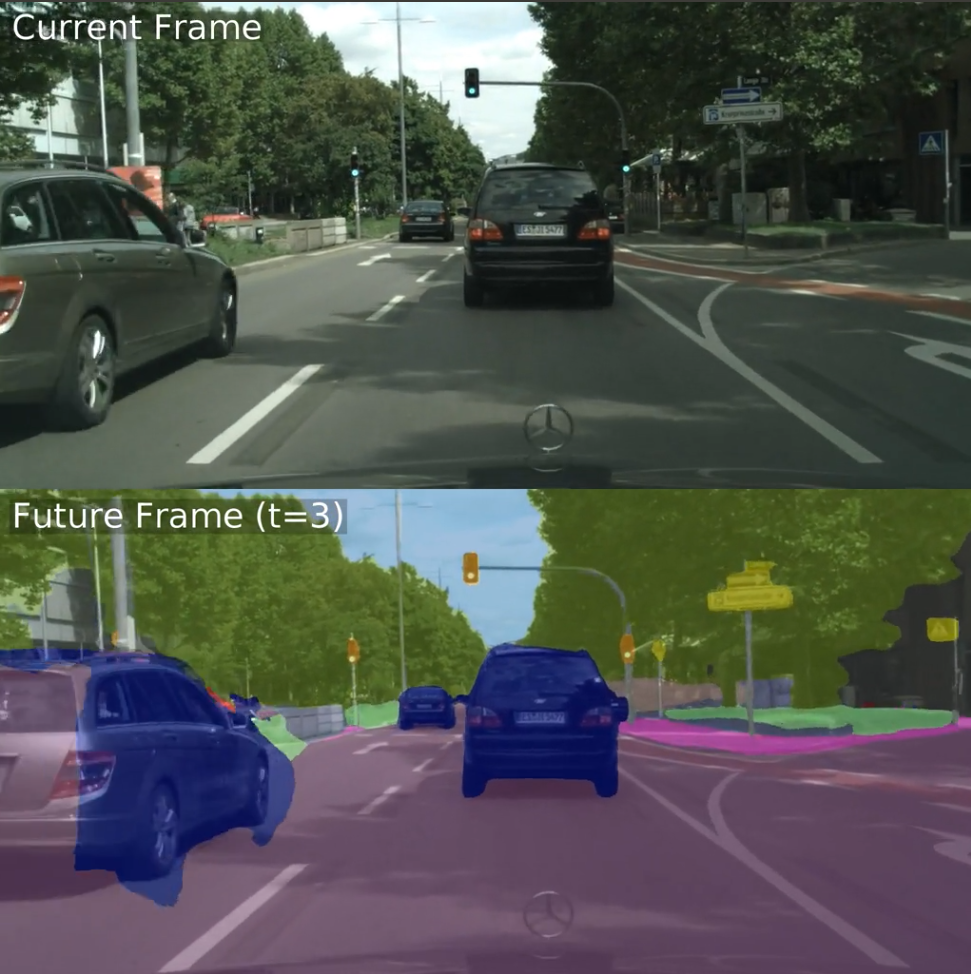}
\includegraphics[width=0.49\linewidth]{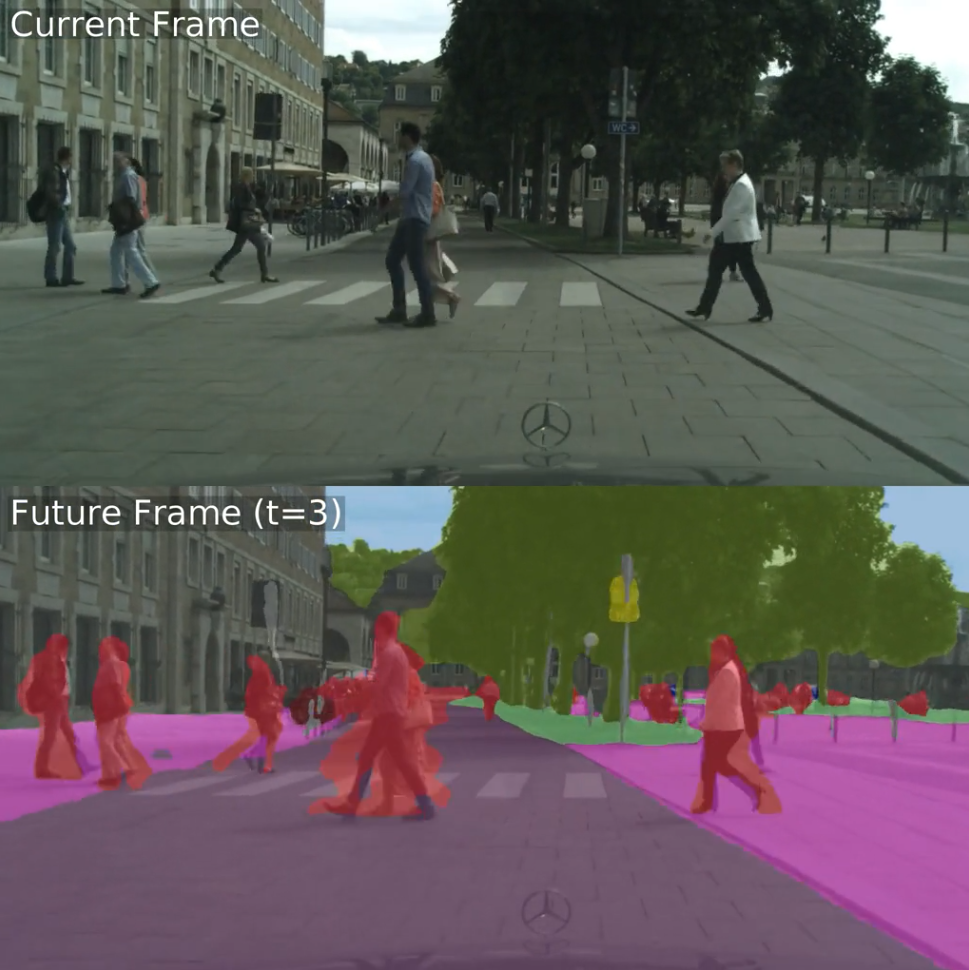}
\caption{We can visualize two distinct failure cases: occlusion and shape degradation. On the left, we our approach struggles reconstruct when objects enter a scene. On the right, our approach can correctly forecast pedestrians, but doesn't maintain their shape well. Future work including inpainting and object consistency may address these issues.} 
\label{fig:failure}
\end{figure}

Finally, in Tab.~\ref{tab:nips}, we compare with the state-of-the-art method on $t=1$ and $t=10$, \cite{Jin_NIPS2017}. For next frame prediction, we demonstrate a new state-of-the-art pushing the margin by $7.1\%$. For $t=10$ prediction, our method performs $1.4\%$ worse overall and only $0.1\%$ worse when controlling for recurrent fine-tuning. It should be noted that the model that performs $52.5\%$ achieves roughly a $0.8\%$ boost due to in-painting. More specifically this occurred when the flow field was occluded and predicted a 0 for flow. 
Thus, we utilized a simple inpainting method which directly copies the current frame segmentation to provide an additional small boost. However, in Fig.~\ref{fig:failure}, we can observe failure cases motivating future work.

In summary, we achieve state-of-the-art performance on short-term ($t=1$, $t=3$) and moving objects ($t=3$, $t=9$) prediction. 
We further find only $\approx$$1\%$ performance degradation on mid-term ($t=10$) prediction, while reducing the number of model parameters by $\approx$$95\%$ and accordingly increasing model efficiency $42x$. 

\subsection{Ablation studies}

In this section, we study the effects of our model from the persepectives of: warp layer, FlowLSTM, time / step size, and auto-regressive vs. single-step. 

\subsubsection{Impact of warp layer} Our main enabling factor for strong performance on next frame and short-term prediction is the warp layer. We can see in Tab.~\ref{tab:warp} how significant the degradation performance is when the warp layer is removed from our system. 
It is such an integral component that our method would not function without it. 
Thus, substituting convolutional layer and concatenation / element-wise addition operations was proven highly ineffective. 
To truly ablate the impact of the warp layer, there would likely need to be a SegPredCNN in addition to these concatenation / element-wise addition operators. 
However, such a change runs contrary to the main motivation for our work being effective with low overhead.

\begin{table}[t!]
\begin{center}
\begin{tabular}{ccc} \toprule Configuration & IoU ($t=3$) & IoU ($t=10$)  \\
    \midrule
       Addition &  $53.8$ & $37.3$ \\
      Concatenation & $54.8$ & $38.6$ \\
       Warp Layer & $\bf66.0$ & $\bf50.0$\\
    \bottomrule
  \end{tabular}
  \end{center}
      \mytabcapspace
 \caption{Impact of warp layer, all configurations use the single-step approach and contain the FlowLSTM.} 
      \label{tab:warp}
            \mytabspace
\end{table}

\begin{table}[t!]
\begin{center}
\begin{tabular}{ccc} \toprule Configuration & IoU ($t=3$)  & IoU ($t=10$)  \\
    \midrule
      No FlowLSTM & $62.3$ & $46.0$ \\
       FlowLSTM  & $\bf 66.0$ & $\bf 50.0$\\
    \bottomrule
  \end{tabular}
  \end{center}
    \mytabcapspace
 \caption{Impact of FlowLSTM, all configurations use the single-step approach and contain the warp layer.} 
  \label{tab:lstm}
      \mytabspace
\end{table}

\subsubsection{Impact of FlowLSTM} 
One novel aspect of our work relative to others in semantic forecasting is the usage of a LSTM. 
 Specifically, although others utilized recurrent fine-tuning to improve their systems with BPTT; our system is the first, to the best of our knowledge, to contain a memory module. 
In Tab.~\ref{tab:lstm}, the effects of FlowLSTM without recurrent fine-tuning are shown, which demonstrates the value of the memory module within LSTM, for both short- and mid-term prediction.

\subsubsection{Impact of time and step size} We can observe in Tabs.~\ref{tab:time}-\ref{tab:step}, the benefit of including more past frames and using a small step size for short-term prediction. Smaller step size worked together with the optical flow network trained on next frame flow. Although notice the diminishing returns going further back in time from $4$ to $8$ past steps. Our method, using the FlowLSTM aggregation, is uniquely able to process frames at a step size of $1$ and a predict a jump of $3$ into the future. Similarly, our best performing mid-term model processes past frames at step size of $3$ to predict a jump of $10$. Another aspect of our approach is the lack of redundant processing, as we only process each frame twice, compared to four times with previous methods (using first-in-first-out concat.). Further, our method can process any number of past frames while maintaining its prediction integrity and run-time efficiency.

\begin{table}[t!]
\begin{center}
\begin{tabular}{ccc} \toprule Time & Frames & IoU ($t=3$)  \\
    \midrule
      $2$ & $15, 16$ & $62.4$ \\
      $4$ & $13, 14, 15, 16$ & $67.0$ \\
      $8$ & $7, 8, ..., 16$ & ${\bf 67.1}$ \\
    \bottomrule
  \end{tabular}
  \end{center}
      \mytabcapspace
    \caption{Impact of time, all configurations use single-step  and contain both the FlowLSTM and warp layer.}
  \label{tab:time}
  \end{table}

\begin{table}[t!]
\begin{center}
\begin{tabular}{ccc} \toprule Step Size & Frames  & IoU ($t=3$)  \\
    \midrule
      $9$ & $7, 16$ & $62.6$\\
      $3$ & $7, 10, 13, 16$ & $66.0$ \\
     $1$ &  $7, 8, ..., 16$ & ${\bf 67.1}$ \\
    \bottomrule
  \end{tabular}
\end{center}
    \mytabcapspace
  \caption{Impact of step size, all configurations use the single-step approach and contain both the FlowLSTM and warp layer.}
  \label{tab:step}
\end{table}

\begin{table}[t!]
\begin{center}
{\setlength{\tabcolsep}{0.9em}
\begin{tabular}{lcc} \toprule Configuration & IoU ($t=3$) & IoU ($t=10$) \\
    \midrule
      Auto-regressive & $64.4$ & $48.7$ \\
            Single-step  & $\bf 66.0$  & $\bf 50.0$ \\
    \bottomrule
  \end{tabular}
}
  \end{center}
      \mytabcapspace
 \caption{Impact of auto-regressive vs. single-step methods for both short-term and mid-term prediction, all configurations contain both the FlowLSTM and warp layer. 
 }
  \label{tab:stepvrecur}
\end{table}

\subsubsection{Impact of auto-regressive vs. single-step} 
In Tab.~\ref{tab:stepvrecur} we show that both auto-regressive and single-step approaches are roughly equivalent in terms of accuracy, with only a slight degradation in favor of the single-step approach. We conjecture that error propagation with $3$ recursive steps is more significant than doing a single large step. 
The overall pros and cons of single-step vs.~auto-regressive depend upon the real-world application deployment. 
For instance, if efficiency is the highest priority in an autonomous vehicle, limiting the processing of redundant frames with a single-step model would be beneficial. 
On the other hand, if flexibility to predict any time step in the future is desired, an auto-regressive approach may be preferred.

%% file: wacv19_con.tex
\section{Conclusion}
In this paper, we posed semantic forecasting in a new way --- decomposing motion and content. 
Through this decomposition into current frame segmentation and future optical flow prediction, we enabled a more compact model. 
This model contained three main components: flow prediction network, feature-flow aggregation LSTM, and an end-to-end warp layer. 
Together these components worked in unison to achieve state-of-the-art performance on short-term and moving objects segmentation prediction.
Additionally, our method reduced the number of parameters by up to $95\%$ and demonstrated a speedup beyond $40x$. 
In turn, our method was designed with low overhead and efficiency in mind, an essential factor for real-world autonomous systems. 
As such, we proposed a lightweight, modular baseline for recurrent flow-guided semantic forecasting.